\pdfoutput=1

\documentclass[11pt]{article}

\usepackage[]{acl}

\usepackage{times}
\usepackage{latexsym}

\usepackage[T1]{fontenc}

\usepackage[utf8]{inputenc}

\usepackage{microtype}

\usepackage{graphicx} 
\graphicspath{{figure/}}
\usepackage{subfigure} 
\usepackage{mathrsfs}
\usepackage{amsmath}
\usepackage{multirow}
\usepackage{bm}
\usepackage{makecell}
\usepackage{color}
\usepackage{amsfonts,amssymb} 
\usepackage{microtype}
\usepackage{kky}
\usepackage{xspace}

\newcommand{\cref}[1]{\S\ref{#1}}

\newcommand{\method}{PHAED\xspace}
\title{A Speaker-aware Parallel Hierarchical Attentive Encoder-Decoder Model for Multi-turn Dialogue Generation}

\author{
  \small Zihao Wang \\ 
  \small Tongji University  \\ 
  \small \texttt{1910658@tongji.edu.cn} \\
  \And
  \small Ming Jiang \\ 
  \small University of Illinois at Urbana-Champaign \\ 
  \small \texttt{mjiang17@illinois.edu} \\ 
  \And
  \small Junli Wang \\ 
  \small Tongji University  \\ 
  \small \texttt{junliwang@tongji.edu.cn} \\ }

\begin{document}
\maketitle
\begin{abstract}
This paper presents a novel open-domain dialogue generation model emphasizing the differentiation of speakers in multi-turn conversations. Differing from prior work that solely relies on the content of conversation history to generate a response, we argue that capturing relative social relations among utterances (i.e., generated by either the same speaker or different persons) benefits the machine capturing fine-grained context information from a conversation history to improve context coherence in the generated response. Given that, we propose a speaker-aware Parallel Hierarchical Attentive Encoder-Decoder (PHAED) model that aims to model each utterance with the awareness of its speaker and contextual associations with the same speaker's previous messages. 
Specifically, in a conversation involving two speakers, we regard the utterances from one speaker as responses and those from the other as queries. After understanding queries via our encoder with inner-query and inter-query encodings, our decoder reuses the hidden states of previously generated responses, instead of reconstructing these
by the encoder, to generate a new response.
Our empirical results show that PHAED outperforms the state-of-the-art in both automatic and human evaluations. Furthermore, our ablation study shows that dialogue models with speaker tokens can generally decrease the possibility of generating non-coherent responses regarding the conversation context.
\end{abstract}

\section{Introduction}

Dialogue generation is a text generation task that aims to automatically generate a reasonable response regarding a conversation history. With the goal of providing AI-based virtual agents to support various services such as personal secretary, companion to humans with an emotional connections and customer services, this task has become a popular research topic in both academia and industry~\citep{Chen2017Survey}. 
Traditional dialogue models are mainly task-oriented~\citep{henderson2013deep, zhao2016towards, madotto2018mem2seq}. To improve the generalization ability of these AI models, recent studies have begun to focus on the development of open-domain conversational agents~\citep{serban2016HRED,xing2018HRAN,chen2019mapping,bao2020plato}.

\begin{figure}[t]
    \centering
    \vspace{-3ex} 
    \resizebox{\columnwidth}{!}{
    \includegraphics[scale=0.70]{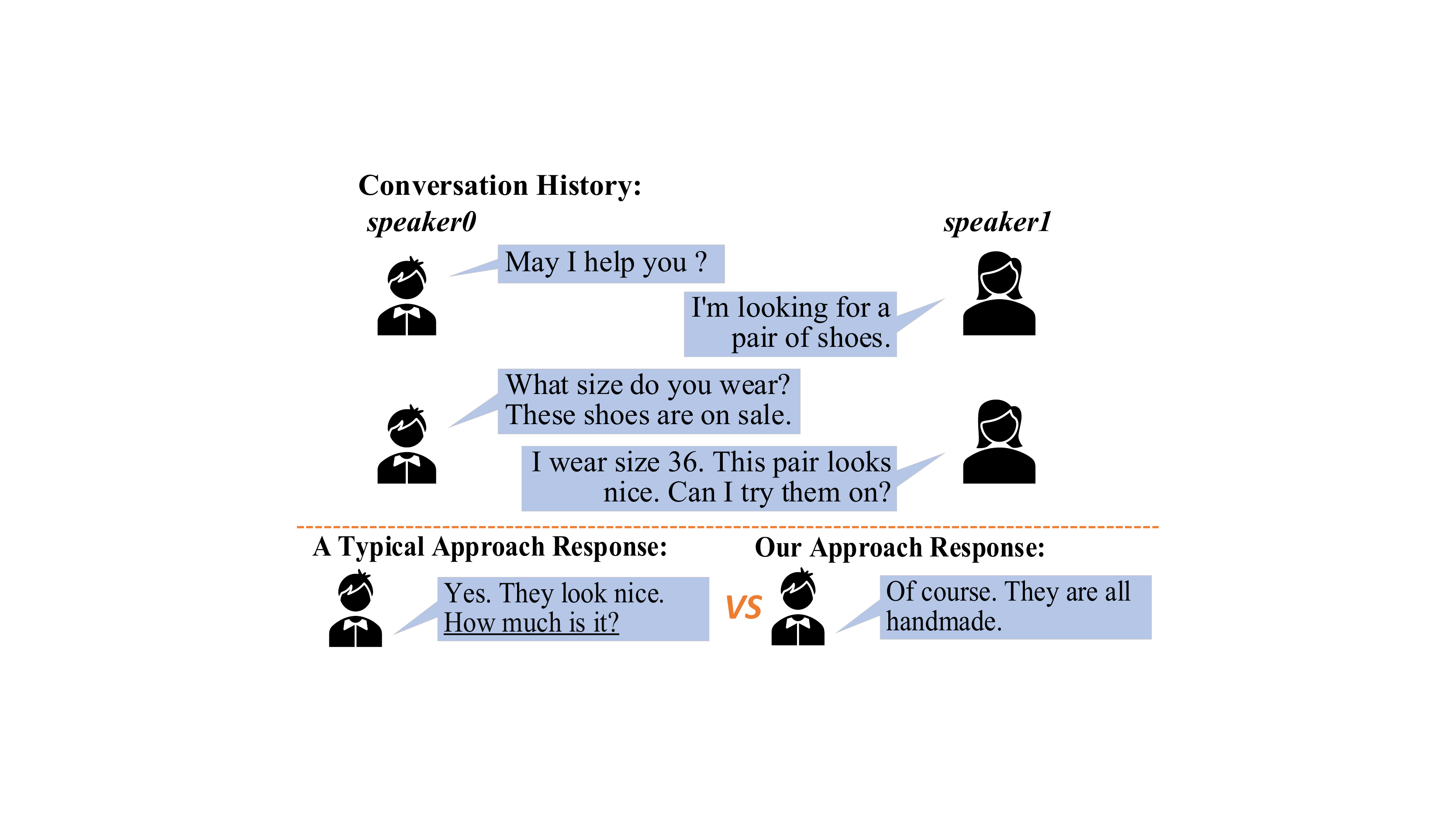}
    }
    \vspace{-3.5ex}   
    \caption{\small A sample of the responses generated by a typical approach (Transformer) and our approach. 
    \emph{speaker0} and \emph{speaker1} are speaker tokens. \underline{Underlined} words indicate the part that does not meet the speaker token.}
    \vspace{-3ex}   
    \label{fig:introduction_sample}
\end{figure}

The common practice of building a dialogue model is to train a sequence-to-sequence model using the conversation history to generate a context-coherent response. Considering the difficulty of understanding the complex scenarios in real life as social interaction, merchandising, and small talk, how to effectively understand conversation history is a critical challenge in the encoding process~\citep{zhao2020AuxiliaryTasks, tian2017useful}. To address this challenge, prior work usually proposes a learning-based model based on the framework of hierarchical recurrent encoder-decoder (HRED)~\citep{serban2016HRED}, where the model contains word-level and utterance-level encoders to understand the conversation history. 

Despite remarkable contributions made by prior work for capturing context information from the conversation history~\citep{serban2017VHRED, tian2017useful, zhang2018DSHRED, xing2018HRAN, zhang2019ReCoSa, zhao2020AuxiliaryTasks}, one major limitation of these studies is that they primarily focus on previous utterances' content but ignore the social relationships between these utterances (i.e., generated by the same or different speaker)~\citep{hovy2021SocialFactors}.
We argue that such missing information is helpful with the machine in learning fine-grained context information by differentiating the content of previous utterances based on their speakers. By losing the chance to get such fine-grained context information, the machine is difficult to capture the latent properties of the speaker (e.g., role) represented by the machine for response generation, and hence, hard to guarantee the context coherence of the generated response. As shown in Figure~\ref{fig:introduction_sample}, we can infer from the conversation history that \emph{speaker0} knows the shoes' price from his message "These shoes are on sale". However, due to the lack of awareness of the speaker information behind utterances, the response generated by an existing dialogue model based on transformer~\citep{vaswani2017transformer} mistakenly asks "How much is it". 

To address the aforementioned limitation, we propose a speaker-aware learning model towards improving multi-turn dialogue coherence via making full use of the conversation history from different speakers. Instead of mixing each utterance with all conversation history, our approach aims to model each utterance with the awareness of its speaker and contextual associations with the same speaker's previous messages. More specifically, to make the model distinguish between queries and responses, we first add different speaker tokens at the beginning of queries and responses. Then, a hierarchical attentive encoder with two-level encodings is proposed to obtain the local and global contextual representations of queries. Finally, the decoder utilizes the turn-level recurrence and cross attention to take advantage of both previous responses and queries for generating the current response. Moreover, we argue that it is unnecessary to re-understand the generated responses since the model must have understood a response before synthesizing it. Therefore, our decoder reuses hidden states of the previously generated responses instead of reconstructing these by the encoder. After considering the speaker roles, we can see from Figure~\ref{fig:introduction_sample} that our approach generates a coherent response with respect to the context of \emph{speaker0}. 

Our main contributions include: 
\begin{itemize}
\item We propose a novel dialogue generation model, PHAED, to generate context-coherent responses in multi-turn dialogues by dealing with utterance information with the awareness of their speakers.
\item By performing experiments on three public datasets, we show that our approach outperforms the state-of-the-art in terms of response coherence and diversity.
\item We conduct a fine-grained analysis of the performance of PHAED, which deepens our understanding of the characteristics of PHAED.
\end{itemize}

\section{Related Work}

As the multi-turn dialogue generation is accordant with the scenarios in daily life, it has gained increasing attention. Besides, since using plain texts has a clear limitation in dialogue generation, it is crucial to making most of the semi-structured data (i.e., containing both textual and auctorial information) in learning neural dialogue models.

Recent work on multi-turn dialogue generation mainly focuses on utterance-aware models towards using conversation history effectively. Early, \citet{serban2016HRED} successfully apply HRED~\citep{sordoni2015HRED} on dialogue generation that models conversation history via a hierarchical recurrent encoder and generates a response by a recurrent decoder. 
Most subsequent studies focus on designing exquisite attention mechanisms to detect the relevant words or utterances for response generation~\citep{tian2017useful, xing2018HRAN, zhang2018DSHRED, zhang2019ReCoSa, zhang2020dialogpt}. \citet{sun2021CoherentDialogueCVAE} and \citet{xu2021DialogGraph} use RNN-based variational auto-encoders to generate responses that are relevant to the content of the conversation history, but they do not take into account the differences between speakers.
\citet{sankar2019effectively} find that transformer-based models have lower test perplexities than recurrent models. 

Regarding speaker-aware models, some methods are proposed on conversational language understanding and generation tasks.
\citet{chi2017speaker} propose speaker role-based contextual model for language understanding and dialogue policy learning. \citet{meng2018multiparty} propose a speaker classification task in multi-party conversation. \citet{ma2019implicit} study the implicit discourse relation identification between different utterances. \citet{liu2020speakerlistener} consider extra fine-grained manual roles (\emph{speaker} or \emph{listener}) of each utterance for multi-turn dialogue generation. 
Besides, there are also some persona-based conversation models~\citep{li2016persona, Olabiyi2018AdversarialPersona, bak2019variational, chan2019modelingwa}, which extract the persona characteristics from the same person's conversations. They require a corpus that has specific identifiers of each particular person. However, for general conversational datasets~\citep{lowe2015Ubuntu, li2017dailydialog}, in different conversations, speakers can only be unified and anonymous as speaker0 and speaker1, rather than labeling all speakers that occurred in the dataset with specific personal identifiers.
In this general setting, \citet{liu2020filling} finds that the response selection model benefits from filling the gap of utterance-aware and speaker-aware representations, and ~\citet{zhao2019effective} propose a speaker-aware generative dialogue model with relative speaker modeling and relative utterance encoders. ~\citet{hovy2021SocialFactors} point out that the reason for the current limitations for NLP applications is the focus on the content of information while ignoring the social factors of the language.

Compared with utterance-aware dialogue generation models, our approach not only focuses on the content but also considers the speaker roles of utterances. Moreover, instead of requiring the extra fine-grained manual labels~\citep{liu2020speakerlistener} or extra conversations with specific personal identifiers~\citep{li2016persona}, we aim at improving multi-turn dialogue coherence in a conversation by the awareness of which utterances from the same speaker. 

\section{Approach}

\begin{figure*}[t]
    \centering
    \vspace{-2ex} 
    \resizebox{\textwidth}{!}{
    \includegraphics[scale=0.58]{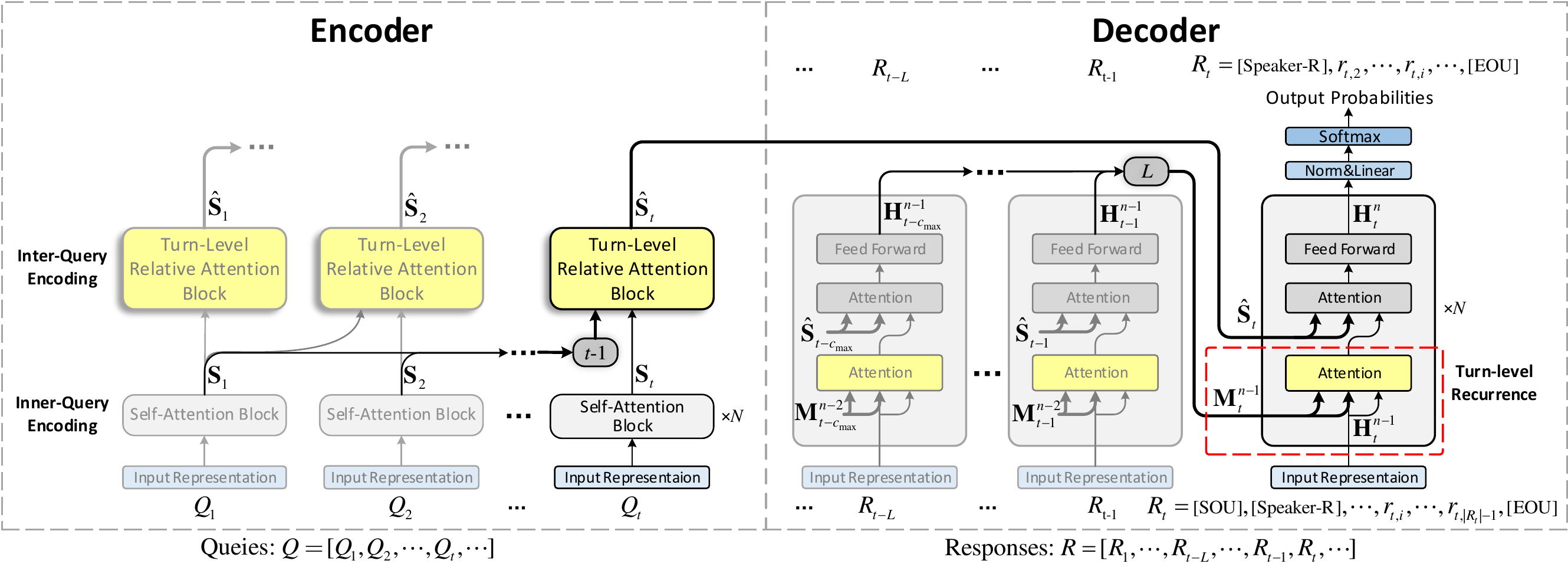}
    }
    \vspace{-4ex}
    \caption{\small The architecture of PHAED. Given a conversation involving a query set and a response set, $\Sbb_{t}$ and $\hat{\Sbb}_{t}$ denote the local and global contextual representations of the $t$-th query $Q_t$. $\Hb_{t}^{n}$ denote hidden states of $t$-th response from $n$-th decoder block. $c_{max}$ denotes memory length of the decoder.
    }
    \vspace{-3ex}
    \label{fig:figure2}
\end{figure*}

Figure~\ref{fig:figure2} provides an overview of our approach. Following the regular dialog flow, we regard utterances in each turn as a query-response pair, and the order of two speakers in a conversation is usually consistent. With the assumption that differentiating the speaker of previous utterances should be helpful with the model's sensibility to the conversation context to generate coherent responses, our goal is to design a multi-turn dialogue generation framework (i.e., PHAED) with the consideration of the speaker role of utterances in the process of generating a context-coherent response in each turn. 

Overall, we will describe PHAED from four aspects:
(1)~We first formalize the problem in \cref{sec:problem formalization}; 
(2)~For modeling multi-turn conversation involving speaker roles, the input representation is designed in \cref{sec:input representation}; 
(3)~Given queries from speaker-Q, the hierarchical attentive encoder constructs local contextual representations (inner-query encoding) and then combines all of them to obtain global contextual representations (inter-query encoding) in \cref{sec:encoder}; 
(4)~After understanding queries, the decoder generates its current response based on the global contextual representations of the queries, the hidden states of its previous responses, and the local context of its partial current response in \cref{sec:decoder}.

\subsection{Problem Formalization}
\label{sec:problem formalization}

Suppose that we have a multi-turn dialogue dataset $\Dcal=\{D^m\}_{m=1}^M$, where $D^m$ is the $m$-th conversation and $M$ is the number of all conversations in the dataset. Each conversation $D^m$ involves two speakers who give queries (i.e., \texttt{[Speaker-Q]}) and responses (i.e., \texttt{[Speaker-R]}) iteratively. Hence we represent each conversation $D^m$ as a sequence of query-response pairs denoted as $D^m = Q^m \otimes R^m$, where $Q^m=(Q_1^m, \dots, Q_{T_m}^m), R^m=(R_1^m, \dots, R_{T_m}^m)$, $T_m$ is the number of pairs. Dialogue modes often adopt an encoder-decoder framework. Here dialogue models aim to generate response $R^m$ given queries $Q^m$ in order. Training criterion is to maximize the conditional log-likelihood, which can be formulated as:
\vspace{-1ex}
\begin{equation}
\small
\sum_{m=1}^M \sum_{t=1}^{T_m} 
         \log P(R_t^m|Q_{\leqslant t}^m,R_{<t}^m;\theta), 
\vspace{-0.5ex}
\end{equation} 
where $Q_{\leqslant t}^m$ refers to all previous queries up to the $t$-th turns, $R_{< t}^m$ denotes the previous responses prior to $R_{t}^m$, and $\theta$ denotes parameter of the model.

\subsection{Input Representation}
\label{sec:input representation}
To distinguish the speaker identities over a multi-turn conversation, we design a novel speaker-aware input representation for words in the query and response utterances. More specifically, we first append two speaker tokens (i.e., \texttt{[Speaker-Q]} and \texttt{[Speaker-R]}) to the beginning of all queries and responses respectively. We then prepend a start-of-utterance token (i.e., \texttt{[SOU]}) and append an end-of-utterance token (i.e., \texttt{[EOU]}) to each utterance. Finally, we add the turn-level and token-level position embeddings to the token embedding as the input representation: 
\vspace{-0.5ex}
\begin{equation}
\label{e_input}
\small 
\begin{split} 
B(q_{t,i}) = E(q_{t,i}) + TE(t) + PE(i)  \\
B(r_{t,i}) = E(r_{t,i}) + TE(t) + PE(i) 
\end{split}  
\end{equation} 
where $q_{t,i}$ ($r_{t,i}$) is the $i$-th word in the $t$-th query (response).
$E(\cdot)$ looks up a token embedding from a embedding matrix. $TE(t)$ is the aligned turn-level embedding indicating the position of the $t$-th utterance. $PE(i)$ is the token-level embedding indicating the position of $q_{t,i}$ ($r_{t,i}$) in the $t$-th utterance. All embeddings are learnable in training.
A detailed visualization example of our input representation structure is provided in appendix~\ref{sec:appendix_input}.

\subsection{Hierarchical Encoder}
\label{sec:encoder} 
We want the encoder to capture and encode all the external information passed to \method. In other words, the encoder is responsible for understanding all queries from other people (i.e., speaker-Q). Since there are multiple queries from the same person and each query has its information, we use two steps to understand all queries. Figure~\ref{fig:figure2} (left) shows the architecture of our encoder.
We first encode each query by self-attention blocks in \emph{Inner-Query Encoding}. Then, we need to combine all information of queries, so we propose \emph{turn-level relative attention} in \emph{Inter-Query Encoding} to consider all queries comprehensively.

\subsubsection{Inner-Query Encoding}
\label{sec:Inner-Query encoder}
To summarize the information from the individual query, we apply a standard $N$-layer Transformer encoder~\citep{vaswani2017transformer} to encode each query. Specifically, we obtain a hidden representation matrix $\Sbb_t^N \in \RR^{|Q_t| \times d_s}$ for all words in the $t$-th query $Q_t$ from the top layer, where $|Q_t|$ is the length of $Q_t$. To simplify the notation, we skip the superscript $N$ hereafter, i.e., $\Sbb_t$. Notably, we adopt the pre-normalization~\citep{bao2020plato2} which has proven effective to stabilize the performance.

\subsubsection{Inter-Query Encoding}
As historical contexts from the query speaker are crucial for understanding the current query, we aim to combine the information from all preceding queries to obtain a global context. To this end, we introduce an inter-query encoding method that obtains a global contextual representation $\hat{\Sbb}_t\in\RR^{|Q_t| \times d_s}$ for $Q_t$ based on a set of the preceding queries denoted as $\Scal_{\leqslant t}=\{\Sbb_1, ..., \Sbb_t\}$, where each element is obtained in~\cref{sec:Inner-Query encoder}. In particular, we propose a \textit{turn-level relative attention} network that extends the relative attention from token-level~\citep{shaw2018Relative} and segment-level~\citep{Zheng2020Translation} to turn-level:  
\begin{equation}
\small
\begin{split}
\label{inter-query}
    \hat{\Sbb}_t &=\text{FFN}(\text{TurnRelAttn}(\Sbb_t, \Scal_{\leqslant t}, \Scal_{\leqslant t})), 
\end{split}
\vspace{-1ex}
\end{equation}
where FFN$(\cdot)$ denotes a feedforward network and TurnRelAttn$(\cdot)$ is our attention network that takes $\Sbb_t$ as the query, and $\Scal_{\leqslant t}$ as the keys and values.

\paragraph{Turn-level relative attention} As the conversation goes on, each query appears in order (from $Q_1$ to $Q_t$). In some cases, $Q_{t}$ focuses on queries that are closer to it. For example, $Q_{t}$ may pay more attention to the closest query $Q_{t-1}$ than other previous queries since $Q_{t-1}$ may contain more relevant information to $Q_{t}$ than others. 

To consider the turn-level relative position among queries in the history, we compute an attention operation from the $t$-th query to the past $p$-th query. Specifically, we first compute the attention's query, key, and value matrices by multiplying the corresponding weight matrices, i.e., $\Wb^Q, \Wb^K, \Wb^V \in \mathbb{R}^{d_s \times d_s}$, and then add the relative position information to the keys and values:
\vspace{-2ex}
\begin{equation}
\small
\begin{split}
    \Qb_t, \Kb_p, \Vb_p &= \Sbb_t \Wb^Q, \Sbb_p \Wb^K, \Sbb_p \Wb^V, ~\forall\Sbb_{p} \in \Scal_{\leqslant t}  \\
     \hat{\Kb}_p  &= \Kb_p + (\Bb^K_{r} \Ib)^\top\\
     \hat{\Vb}_p  &= \Vb_p + (\Bb^V_{r} \Ib)^\top\\
     r &= \min( t - p, r_\text{max}) 
\end{split}
\end{equation} 
where $r$ measures the relative position between $t$ and $p$ up to a pre-defined maximum number $r_\text{max}$, $\Bb^K,~\Bb^V \in \RR^{d_s \times r_\text{max}}$ are two learnable matrices that captures the relative position information for the attention's keys and values, and 
$\Ib\in\RR^{1\times |Q_p|}$ 
is an all-one row vector. Here we take the $r$-th column vector $\Bb_r^K$ from $\Bb^K$ and copy it $|Q_p|$ times across columns by multiplying it with $\Ib$. Similar operations apply to $\Bb_r^V$.

We then compute the attention matrix $\Ab_{t\rightarrow p}\in \RR^{|Q_t|\times |Q_p|}$, and obtain a global contextual representation $\hat{\Sbb}_t$ by a weighted sum over the values of all preceding queries, followed by a residual connection and a feedforward network as follows:
\begin{equation}
\small
\label{eq:Atp}
\begin{split}
    \Ab_{t\rightarrow p} &= \text{softmax}\left(\frac{\Qb_{t}  \hat{\Kb}_{p}^\top}{\sqrt{d_s}}\right),  \\ 
    \hat{\Sbb}_{t} &= \text{FFN} \left(\Sbb_{t} + \sum_{p=1}^t \Ab_{t\rightarrow p} \hat{\Vb}_{p} \right) \\
\end{split} 
\end{equation}

\subsection{Decoder with Turn-level Recurrence}
\label{sec:decoder}
PHAED understands other people's queries through the encoder, but it still needs to consider its previously generated responses from Speaker-R for generating its current response. Our idea is to store the hidden states of its previous responses as memory and reuse the memory as the information of these responses to generate its current response. For this purpose, as shown in figure~\ref{fig:figure2} (right), we take the Transformer-xl~\citep{dai2019transformer-xl, Zheng2020Translation} with cross attention as the decoder.

\paragraph{Turn-level Recurrence.} 
In~\citet{dai2019transformer-xl}, the decoder is consists of $N$ Transformer layers, where each layer augments the attention's keys and values from the previous layer by caching the hidden states of a fixed length of previous words in the memory. This caching design allows the decoder to access a longer context in the memory. Similarly, we aim to extend the context from preceding responses for the decoder and encourage the decoder to capture the turn-level relationship between responses. To this end, we cache the hidden states of words from at most $c_\text{max}$ previous responses and concatenate them along the length dimension as the memory $\Mb_t^{n-1}$ from the $(n-1)$-th layer:
\vspace{-1ex}
\begin{equation}
\small
\begin{split}
    c &= \max (t-c_\text{max}, 1), \\
    \Mb_t^{n-1} &= [\Hb_{c}^{n-1}\circ \dots\circ \Hb_{t-2}^{n-1} \circ \Hb_{t-1}^{n-1}],\\
\end{split}
\end{equation}
where $\Hb_c^{n-1}\in \RR^{m_c \times d_s}$ in the memory denotes the hidden states of the $c$-th response of word size $m_c$ from the $(n-1)$-th transformer layer. Similar to \citet{dai2019transformer-xl}, we truncate the gradient from the memory, augment the attention's keys and values with the memory, and obtain the next layer's hidden states $\Hb_t^n$ by a transformer layer.
\vspace{-1ex}
\begin{equation}
\small
\begin{split}
    \widetilde{\Hb}_t^{n-1} &= [\text{SG}(\Mb_t^{n-1}) \circ \Hb_t^{n-1}], \\
    \hat{\Hb}_t^{n}
    &=\text{Attention}(\Hb_t^{n-1},\widetilde{\Hb}_t^{n-1},\widetilde{\Hb}_t^{n-1}), \\
    \Hb_t^{n}
    &=\text{FFN}(\text{Attention}(\hat{\Hb}_t^{n},\hat{\Sbb}_t,\hat{\Sbb}_t))
\end{split}
\end{equation}
where $\text{SG}(\cdot)$ denotes stop-gradient and the Transformer-layer takes the query, key, and value matrices for cross-attention followed by a feedforward network.

Finally, we obtain the output probabilities of $R_t$ by a linear layer and a softmax layer on the word representations $\Hb_t^N$ from the top decoder layer:
\vspace{-1ex}
\begin{equation}
\small
\begin{split} 
P(R_t|R_{<t},Q_{\leqslant t};\theta)
    &= \prod_{i=1}^{|R_{t}|} P(r_{t,i}|r_{t,<i},R_{<t}, Q_{ \leqslant t}; \theta)\\
    &= \prod_{i=1}^{|R_{t}|} \text{softmax}( \Hb_{t,i}^N \Wb^{O})
\end{split}
\end{equation}
where $\Wb^{O}\in \RR^{d_s\times |V|}$ is a linear project matrix for a vocabulary of size $|V|$. $\Hb_{t,i}^N$ denotes the $i$-th row vector from $\Hb_t^N$ and $|R_t|$ is the length of $R_t$.

\section{Experiments}
We conduct experiments on three datasets and compare PHAED with the state-of-the-art based on both automatic and human evaluations. 
Besides, we further conduct a fine-grained analysis to deepen our understanding of the characteristics of PHAED. 

\subsection{Experimental Setup}

\subsubsection{Datasets} 
Three popularly used benchmark datasets for open-domain multi-turn dialogue generation are adopted, which include: (1) DailyDialog, (2) PersonaChat, and (3) Ubuntu v2.0. Conversations in all datasets involve two participants.
Dialogues in \textbf{DailyDialog}~\citep{li2017dailydialog} cover various topics about our daily life such as social activities and school life. 
\textbf{PersonaChat}~\citep{zhang2018personalizing} is a knowledge grounded dataset that contains dialogues and speaker profile information. Following the standard practice of prior work~\citep{zhao2020AuxiliaryTasks}, we append the profile to the conversation history. \textbf{Ubuntu v2.0}~\citep{lowe2015Ubuntu}
is a large multi-turn dialogue corpus extracted from Ubuntu question-answering forum. We truncate the utterances with more than 50 tokens, and the truncated utterances with abnormal endings are corrected. Table~\ref{tab:dataset} provides a detailed statistics of each dataset. 
\begin{table}[h]
    \centering
    \scriptsize 
    \vspace{-5ex}
    \resizebox{\columnwidth}{!}{
    \begin{tabular}{llll}
        \hline
        
        \hline
        & DailyDialog & PersonaChat & Ubuntu \\
        \hline
         \hspace{-2ex}train dialogues & 11,118  & 8,939 & 1,000,000 \\
         \hspace{-2ex}valid dialogues & 1,000  & 1,000 & 19,560\\
         \hspace{-2ex}test dialogues & 1,000  & 968  & 18,920\\
         \hspace{-2ex}avg. utterances per dialogue & 7.9  & 14.8 & 4.9 \\
         \hspace{-2ex}avg. tokens per utterance & 14.6 & 12.9 & 16.2\\
        \hline
        
        \hline
    \end{tabular}}
    \vspace{-3ex}
    \caption{\small Statistics of three datasets.}
    \label{tab:dataset}
    \vspace{-5ex}
\end{table}
\subsubsection{Compared Methods}
We select utterance-aware and speaker-aware state-of-the-art methods as baselines and train all models on the same preprocessed data \emph{with speaker tokens added}: (1)~\textbf{HRAN}: A HRED equipped with hierarchical attention based on the utterance-level and the word-level presentations~\citep{xing2018HRAN}; (2)~\textbf{DSHRED}: A HRED equipped with the static and dynamic attention~\citep{zhang2018DSHRED}; (3)~\textbf{SpkHRED}: 
A recent speaker-aware HRED with relative speaker modeling and relative utterance encoders~\citep{zhao2019effective}; 
(4)~\textbf{Transformer}: Under the encoder-decoder framework for dialogue generation, the most simple but natural idea is to directly use the Transformer~\citep{vaswani2017transformer} to encode all the previous utterances and then decode the representations to generate a response; (5)~\textbf{ReCoSa}: A hierarchical transformer-based model for detecting the relevant contexts~\citep{zhang2019ReCoSa}; and (6)~\textbf{DialoGPT}: Following ~\citet{zhang2020dialogpt}, we train a multi-turn dialogue generation model from scratch on the basis of the GPT-2~\citep{radford2019gpt2}. 

\subsubsection{Implementation Details}
The dimension $d_s$ of hidden states is set to be 512 in HRAN, DSHRED, Transformer($N$=6), ReCoSa($N$=6), PHAED($N$=4) and PHAED($N$=6). $N$ denotes the number of layers. Except for DialoGPT, the head number of all transformer-based models is 8. For DialoGPT($N$=6 and $N$=12), we use the small GPT-2 architecture with $d_s$ 768 and head number 12. 
We increase the $d_s$ to 560 in PHAED($N$=12) to make it has the same parameter size as DialoGPT($N$=12). Greedy search is taken as the decoding strategy. Adam optimizer~\citep{kingma2015adam} with an initial learning rate of 0.005 is utilized for training, and the batch size is 32. We train and evaluate each model on a Tesla P100 card or V100 card with PyTorch.

\subsubsection{Evaluation Measures}
\paragraph{Automatic evaluation} We evaluate PHAED and baselines based on \emph{coherence} and \emph{diversity} metrics. For coherence evaluation, we adopt \textbf{Perplexity}~\citep{xing2018HRAN}, \textbf{BLUE-n} for n-grams ($n$=1,2,3,4)~\citep{tian2017useful}, and three embedding-based metrics~\citep{serban2017VHRED}. Embedding-based metrics include Average(\textbf{Avg}), Extrema(\textbf{Ext}), and Greedy(\textbf{Gre}), using pre-trained Google news word embedding~\citep{Mikolov2013Word2Vec}. 
For diversity evaluation, we use \textbf{Distinct-1} and \textbf{Distinct-2}, which calculate the ratio of unique unigrams and bigrams, respectively. 
\emph{Notably}, since the probability of the speaker token (\texttt{[Speaker-R]}) is much higher than other tokens, the Perplexity with speaker token probability will be much lower than the Perplexity without that. However, we focus on the other tokens in response, so \emph{we do not take the probability of the speaker token into account when calculating Perplexity and remove the speaker token from the generated responses before using other metrics for evaluation.}

\paragraph{Human evaluation} 

We further conduct a manual evaluation to explicitly examine the quality of dialogue models based on human judgments. Following prior work~\citep{xu2021DialogGraph,cai2020Manipulation}, we randomly select 100 examples containing conversation history and responses generated by baselines and PHAED as testing examples. Based on such testing data, we recruit three human annotators with good English skills to score the response quality on a scale of [0, 1, 2] from four aspects, which includes: (1) \textbf{Fluency} in terms of the smoothness of response and the correctness of grammar, (2) \textbf{Coherence} indicating whether the response is coherent with conversation history, (3) \textbf{Informativeness} that focuses on the amount of information contained in the response, and (4) \textbf{Overall} that stands for the general evaluation, where 0, 1, and 2 indicate bad, good, and perfect responses, respectively.

\begin{table*}[t] 
\centering
\vspace{-0ex}
\resizebox{\textwidth}{!}{
\begin{tabular}{c|c|c|llll|lll|ll|c}
    \hline
    
    \hline
    \multirow{2}{*}{\textbf{Dataset}}& \multirow{2}{*}{\textbf{Model}} & \multicolumn{8}{c|}{\textbf{Coherence}} & \multicolumn{2}{c|}{\textbf{Diversity}} & \multirow{2}{*}{\bm{$|\theta|$}} \\
    \cline{3-12} 
    & & \textbf{Perplexity}  $\downarrow$ 
    & \multicolumn{4}{c|}{\textbf{BLEU-1 / 2 / 3 / 4} $\uparrow$} 
    & \multicolumn{3}{c|}{\textbf{Avg / Ext / Gre} $\uparrow$} 
    & \multicolumn{2}{c|}{\textbf{Distinct-1 / 2} $\uparrow$} & \\
    \hline 
    
    \hline 
    \multirow{10}{*}{DailyDialog} & HRAN & 31.04
        &  19.10 & 8.511 & 4.670 & 2.790
        &  63.90 & 38.86 & 45.52
        &  1.732 & 8.700 
        & 32.6M \\
     & DSHRED & 31.71
        & 18.80 & 8.351 & 4.547 & 2.686
        & 63.64 & 38.98 & 45.01
        & 1.492 & 7.606
        & 34.2M \\
    & SpkHRED & 34.22
        & 19.21	& 8.421 & 4.485 & 2.533
        & 63.84 & 38.51 & 45.11
        & 1.052 & 4.510
        & 40.2M \\
    & \textbf{PHAED($\bm{N}$=4)} & \textbf{25.67} 
        & \textbf{19.24} & \textbf{9.200} & \textbf{5.444} & \textbf{3.517}
        & \textbf{64.24} & \textbf{39.47} & \textbf{46.30} 
        & \textbf{2.633} & \textbf{13.80}
        & 42.7M \\
    \cline{2-13}
    \cline{2-13}
     & Transformer($N$=6) & 27.34
        & 17.72 & 7.205 & 3.722 & 2.092
        & 62.85 & 37.88 & 44.58
        & 2.398 & 11.85
        & 50.2M \\
     & ReCoSa($N$=6) & 25.34 
        & 17.77 & 7.186 & 3.684 & 2.100
        & 62.90 & 37.42 & 44.89 
        & 2.481 & 12.39
        & 68.5M \\
     & DialoGPT($N$=6) & 29.93
        & 17.69 & 8.528 & 5.214 & 3.564
        & 64.12 & \textbf{40.24} & 45.98 
        & 1.905 & 12.37
        & 57.2M \\
     & \textbf{PHAED($\bm{N}$=6)} & \textbf{24.45}
        & \textbf{19.02} & \textbf{9.174} & \textbf{5.508} & \textbf{3.602}
        & \textbf{64.42} & 39.82 & \textbf{46.34} 
        & \textbf{2.932} & \textbf{15.58} 
        & 53.8M \\
    \cline{2-13}
    
    \cline{2-13}
     & DialoGPT($N$=12) & 27.89
        & 18.54 & 9.432 & 6.077 & 4.382
        & 64.86 & \textbf{40.70} & 46.89
        & 2.109 & 14.00
        & 99.7M \\
        & \textbf{PHAED($\bm{N}$=12)} & \textbf{23.71} 
        & \textbf{20.47} & \textbf{10.61} & \textbf{7.089} & \textbf{5.326}
        & \textbf{64.91} &	39.96 & \textbf{47.34} 
        & \textbf{3.639} & \textbf{20.19} 
        & 99.3M \\
    \hline
    
    \hline
    \multirow{8}{*}{PersonaChat} & HRAN & 36.59 
        & 21.06 & 10.22 & 5.292 & 2.779
        & 62.18 & 38.21 & 43.48 
        & 0.2396 & 0.9633
        & 34.1M \\
     & DSHRED & 36.96 
        & \textbf{21.69} & 10.46 & 5.401 & 2.841
        & 62.58 & 38.56 & 44.17 
        & 0.2718 & 1.357 
        &  34.8M \\
     & SpkHRED & 38.21 
        & 21.19 & 10.08 & 5.088 & 2.580
        & 61.95 & 37.97 & 43.37 
        & 0.1856 & 0.6902 
        &  40.7M \\
     & \textbf{PHAED($\bm{N}$=4)} & \textbf{33.13} 
        & 21.48 & \textbf{10.51} & \textbf{5.603} & \textbf{3.101} 
        & \textbf{63.60} & \textbf{39.75} & \textbf{45.16} 
        & \textbf{0.5401} & \textbf{2.580}
        & 43.4M\\
    \cline{2-13}
    \cline{2-13}
     & Transformer($N$=6)& 37.09 
        & 19.55 & 8.807 & 4.027 & 1.959
        & 59.84 & 38.05 & 40.19
        & 0.1443 & 0.4177 
        & 51.0M \\
     & ReCoSa($N$=6) & 33.88
        & 21.04 & 9.925 & 4.973 & 2.563
        & 62.64 & 37.40 & 44.54
        & 0.4417 & 1.703 
        & 69.4M \\
     & DialoGPT($N$=6) & 32.91 
        & 20.89 & 10.19 & 5.369 & 2.958
        & 62.69 & 39.38 & 44.43 
        & \textbf{0.5126} & 2.298 
        & 57.7M \\
    & \textbf{PHAED($\bm{N}$=6)} & \textbf{32.62} 
        & \textbf{21.93} & \textbf{10.66} & \textbf{5.595} & \textbf{3.026}
        & \textbf{63.09} & \textbf{39.48} & \textbf{44.98} 
        & 0.4996 & \textbf{2.453}
        & 54.4M \\
    \hline
    
    \hline
    \multirow{4}{*}{Ubuntu} 
     & Transformer($N$=6)& 42.74
        & 11.51 & 3.460 & 1.315 & 0.5546
        & \textbf{62.35} & 34.08 & 43.64
        & 0.08594 &  0.3306
        & 82.9M \\
     & ReCoSa($N$=6) & 36.40
        & 12.74 & 4.750 & 2.183 & 1.150
        & 59.39 & 34.65 & 43.05
        & 0.3970 & 3.441
        & 83.8M \\
     & DialoGPT($N$=6) & 29.98
        & 13.06 & 5.077 & 2.425 & 1.299
        & 59.77 & 35.00 & 44.02
        & 0.6783 & 5.425
        &  81.7M\\
    & \textbf{PHAED($\bm{N}$=6)} & \textbf{28.71} 
        & \textbf{13.15} & \textbf{5.676} & \textbf{2.508} & \textbf{1.449}
        & 60.43 & \textbf{35.72} & \textbf{44.29} 
        & \textbf{0.7328} & \textbf{5.735}
        & 86.4M \\
    \hline
    
    \hline
\end{tabular}}
\vspace{-1.5ex} 
\caption{\small Automatic evaluation results (\%) on three datasets. \emph{Notably}, we train all models on the same preprocessed data with speaker tokens added and evaluate the test results with the speaker tokens removed. $N$ denotes the number of stacked blocks. \bm{$|\theta|$} denotes the parameter size. Memory length $c_{max}$ of PHAED is 1. The best results in each metric are highlighted with \textbf{bold}. ``$\uparrow$'' means higher is better, and ``$\downarrow$'' means lower is better. 
}
\vspace{-2ex}
\label{AutEva}
\end{table*}

\subsection{Evaluation Results} 
Considering the influence of the parameter size, we compare PHAED($N$=4) with RNN-based baselines and PHAED($N$=6) with Transformer-based baselines.
Table~\ref{AutEva} shows the automatic evaluation results, where we observe that PHAED outperforms baselines in three datasets. For PHAED with different $N$, a small increase in $N$ (from 4 to 6) makes PHAED perform better on Perplexity, and with substantial amplification of $N$ (from 4 to 12 or 6 to 12), PHAED achieves better performance on all metrics. Taking the results on DailyDialog as an example, the metrics scores of the PHAED($N$=4 and $N$=6) are better than other baselines overall, and when the models' parameter sizes are the same, PHAED($N$ =12) outperforms DialoGPT($N$=12) on most metrics. Therefore, we demonstrate that PHAED performs better than baselines on automatic evaluation and generates high-quality responses. 
Moreover, with respect to the lower value of Perplexity scores achieved by PHAED with larger $N$, we can infer that stacking more blocks benefits PHAED increasing the possibility of generating coherent and diverse responses based on the conversation history.

\begin{table}[t]  \footnotesize \
\centering
\begin{tabular}{lccccc}
    \hline
    \textbf{Model} &\textbf{Flu.} & \textbf{Coh.} & \textbf{Inf.} & \textbf{Overall}\\
    \hline 
    HRAN  
        & 1.09
        & 0.94
        & 0.89
        & 0.86  \\ 
    DSHRED  
        & 0.90
        & 0.78
        & 0.73
        & 0.69 \\
    SpkHRED %
        & 0.89
        & 0.77
        & 0.69
        & 0.66 \\
    Transformer %
        & 0.97 
        & 0.84
        & 0.92
        & 0.78 \\
    ReCoSa  %
        & 1.00
        & 0.84
        & 0.86
        & 0.77 \\
    DialoGPT %
        & 1.15%
        & 0.97%
        & 0.92
        & 0.93\\
    PHAED 
        & \textbf{1.28}
        & \textbf{1.19}
        & \textbf{1.21}
        & \textbf{1.14}\\
    \hline
\end{tabular}
\vspace{-1.5ex}
\caption{\small Human evaluation results.}
\vspace{-4ex}
\label{HumEva}
\end{table}

We carry out the human evaluation on the DailyDialog that contains a wide variety of high-quality conversations from daily life~\citep{cai2020Manipulation}. The human evaluation results are shown in Table~\ref{HumEva}. The average kappa score~\citep{fleiss1971measuring} is 41.68, which indicates the moderate agreement of the three annotators. From the results, we can see that PHAED achieves better performance on all the metrics than baselines, which indicates that PHAED is preferred by humans. 
The coherence assessments indicate that our responses are coherent with the context. Besides, compared with the baseline methods, the scores of fluency and informativeness are also high, revealing that our approach tends to generate more fluent and informative responses.
 
\begin{table*}[t] \tiny
\centering
\resizebox{\textwidth}{!}{
\begin{tabular}{l|c|llll|lll|ll}
    \hline
    \multirow{2}{*}{\textbf{Model Variant}} &  \multicolumn{8}{c|}{\textbf{Coherence}} &
    \multicolumn{2}{c}{\textbf{Diversity}}
    \\
    \cline{2-11}
    & \textbf{Perplexity} $\downarrow$
    & \multicolumn{4}{c|}{\textbf{BLEU-1 / 2 / 3 / 4} $\uparrow$}
    & \multicolumn{3}{c|}{\textbf{Avg / Ext / Gre} $\uparrow$ }
    & \multicolumn{2}{c}{\textbf{Distinct-1 / 2} $\uparrow$} \\
    
    \hline
    PHAED($N$=4) & \textbf{25.67} 
        & \textbf{19.24} & \textbf{9.200} & \textbf{5.444} & \textbf{3.517}
        & \textbf{64.24} & 39.47 & \textbf{46.30}
        & \textbf{2.633} & \textbf{13.80} \\ 
        
      \textsl{~~~~w/o speaker tokens} & 25.77 
        & 18.05 & 8.416 & 4.855 & 3.045
        & 63.86 & \textbf{39.57} & 45.58  
        & 2.364 & 12.73 \\
      \textsl{~~~~w/o aligned turn embedding} & 26.03 
        & 18.20 & 8.154 & 4.461 & 2.636
        & 63.73 & 39.28 & 45.49 
        & 2.339 & 12.40 \\
      \textsl{~~~~w/o turn-level relative attention} & 26.15 
        & 17.37 & 7.905 & 4.492 & 2.772
        & 62.87 & 38.88 & 44.81
        & 2.335 & 12.08 \\
      \textsl{~~~~w/o turn-level recurrence} & 25.99
        & 18.42 & 8.264 & 4.626 & 2.815
        & 63.72 & 39.27 & 45.72 
        & 2.249	& 11.70 \\
    \hline 
    Transformer($N$=6) & \textbf{27.34} 
        & \textbf{17.72} & \textbf{7.205} & \textbf{3.722} & \textbf{2.092}
        & \textbf{62.85} & \textbf{37.88} & \textbf{44.58} 
        & \textbf{2.398} & \textbf{11.85} \\
    \textsl{~~~~w/o speaker tokens} & 27.93 
        & 17.48 &  7.078 & 3.620 & 1.966
        & 62.51 & 37.78 & 44.44 
        & 1.967 & 9.483 \\
    \hline
    DialoGPT($N$=6) & \textbf{29.93}
        & 17.69 & \textbf{8.528} & \textbf{5.214} & \textbf{3.564}
        & \textbf{64.12} & \textbf{40.24} & \textbf{45.98} 
        & \textbf{1.905} & \textbf{12.37} \\
        
    \textsl{~~~~w/o speaker tokens} & 30.86
        & \textbf{17.71} & 8.474 & 5.124 & 3.458
        & 63.91 & 39.90 & 45.78
        & 1.869 & 11.96 \\
    \hline
\end{tabular}}
\vspace{-1.5ex}
\caption{\small Ablation Study results(\%). We also evaluate test results with the speaker tokens removed. }
\vspace{-2ex}
\label{ablation}
\end{table*}

\subsection{Ablation Study}
To provide a fine-grained analysis of the contribution of each component in PHAED (i.e., speaker tokens and aligned turn embedding in input representations, turn-level relative attention in encoder, and turn-level recurrence in decoder), we conduct an ablation study. Table~\ref{ablation} shows our results. The ablation models without speaker tokens show deteriorations in the majority of metrics such as perplexity, suggesting that adding speaker tokens in the input representations benefits both PHAED and other dialogue models generating coherent and diverse responses regarding the conversation context. 
Without aligned turn embedding that encodes the order of utterances, PHAED achieves a decreasing performance in all metrics. Meanwhile, by removing turn-level relative attention block or memory relative attention, the performance also obviously decreases in all metrics. Therefore, it is critical to consider the utterance-level positional information and the contextual information of both queries and responses. Besides, PHAED without speaker tokens shows less deterioration than the other ablation models of PHAED in most metrics, suggesting that the components we proposed in PHAED play more important roles than the input representations with speaker tokens.

\begin{figure*}[t]
    \centering 
    \resizebox{\textwidth}{!}{
    \includegraphics[scale=0.3]{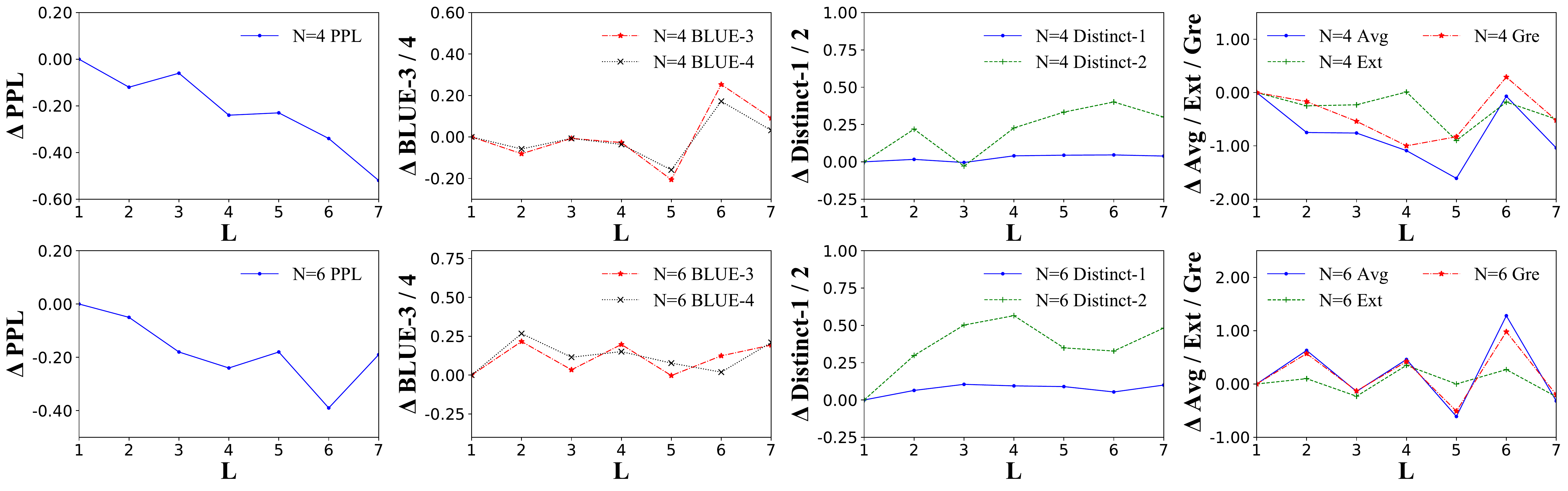}
    }
    \vspace{-4ex}
    \caption{\small The impact of memory length $L=c_{max}$ on the performance of PHAED on PersonaChat. The range of $L$ is from 1 to 7. $\bm{\Delta}$\textbf{metric} represents the change in the score of the metric. Perplexity is abbreviated as \textbf{PPL}. }
    \vspace{-3ex}
    \label{fig:figure4}
\end{figure*}

By looking into the PHAED structure more in-depth, 
we further explore the impact of decoder memory length $L=c_{max}$ (i.e., number of previously machine-generated responses cached in memory) on PHAED's performances to identify the optimal value of $L$. According to \citet{dai2019transformer-xl}, the dependency length of the turn-level recurrence is the sum of the lengths of prior $N \cdot L$ responses. However, is $L$ the bigger, the better? To answer this question, we re-train PHAED with different $L$. The values in the Figure~\ref{fig:figure4} represent the difference between the results of PHAED($L$=1) (Table~\ref{AutEva}) and the results of PHAED($L\ge1$). Overall, with an increase of $L$, PHAED obtains a better perplexity, but there is only a small change on each metric score. 
Considering PHAED with a large memory length costs high computing resources, we empirically set the value of $L$ as 3 in PHAED($N$=4) and 2 in PHAED($N$=6).

\subsection{Case Study}
\begin{figure}[t]
    \centering 
    \includegraphics[scale=0.25]{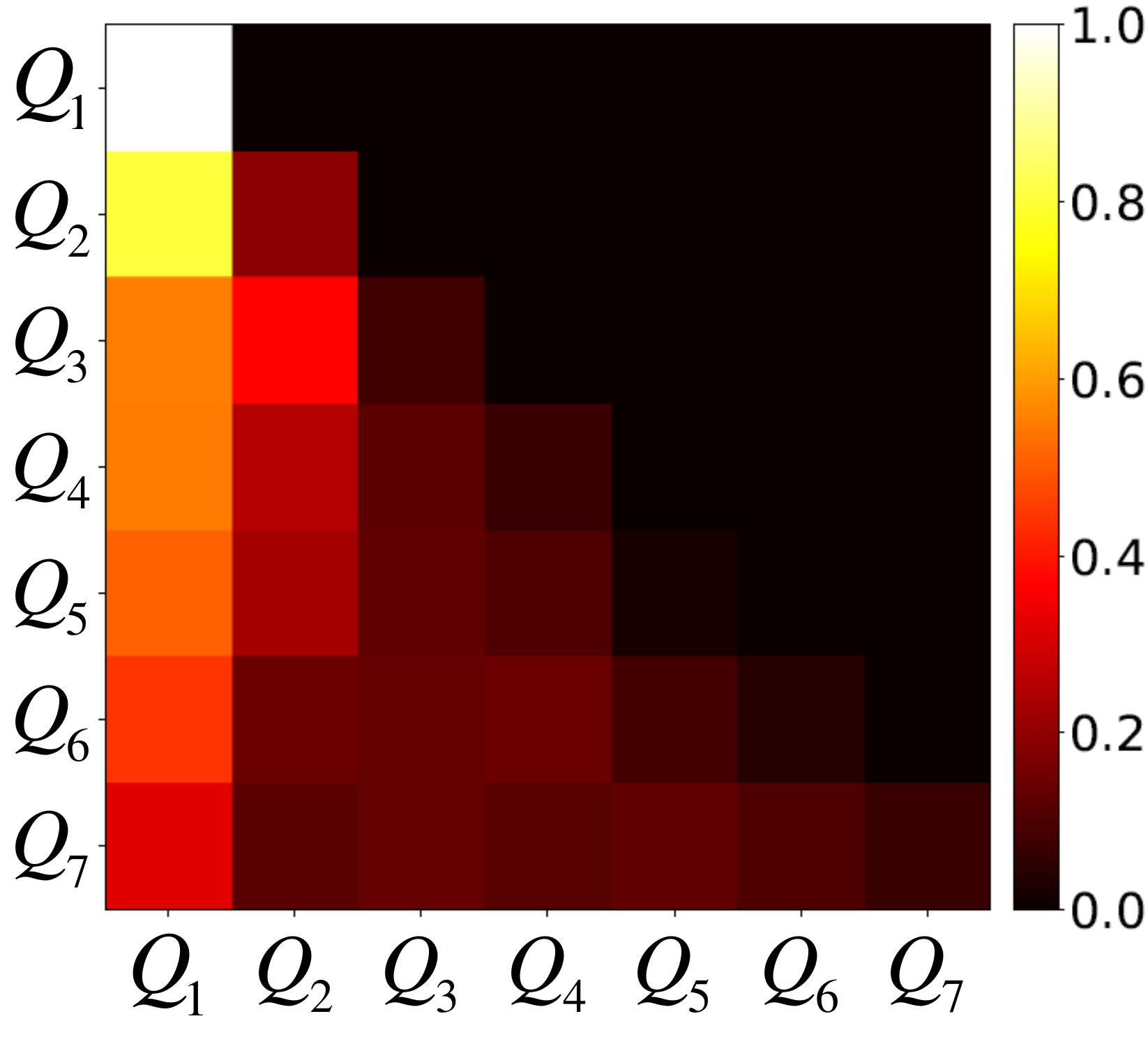}
    \caption{\small Visualization of weights from query-to-query. The total weights of each row are 1.}
    \vspace{-3ex}
    \label{fig:attn_q2q}
\end{figure}
We would like to know what PHAED has learned from the conversation history. We visualize the query-to-query weights of a conversation based on \emph{turn-level relative attention} of PHAED($N$=6). Formally, the weight of the $t$-th query attending to the $p$-th query is computed by $ \alpha_{t}^{p} = \frac{1}{|Q_t|}\text{sum}(\Ab_{t\rightarrow p})$, where $\Ab_{t\rightarrow p}$ is defined by Eq.(\ref{eq:Atp}), and $\text{sum}(\cdot)$ gets the sum of all elements of the input matrix. 
As shown in Figure~\ref{fig:attn_q2q}, two findings are also common in other conversations: the first query $Q_1$ (first column), which contains the major topic of a conversation, seems to be a terrifically useful context for subsequent queries; since representations of current query can be passed by the residual connection, the turn-level attention seems to care less about the current query (diagonal queries). 
Besides, two dialogue examples from the DailyDialog test results are provided in Appendix~\ref{sec:appendix_case_study}.

\section{Conclusion and Future Work}
We have presented a novel learning model called PHAED for multi-turn dialogue generation by utilizing utterance relations based on their speakers to capture fine-grained conversation context information. 
The presented experiments with three benchmark datasets have shown that PHAED outperforms state-of-the-art by improving the context coherence of responses. Moreover, we find that PHAED learns more from utterances containing high-level topic information of a conversation history than other utterances. 

In the future, we will extend PHAED in multi-party conversation, where the encoder is responsible for understanding the utterances of all other speakers (i.e.,~\texttt{Speaker1-Q},~\texttt{Speaker2-Q}, \texttt{Speaker3-Q}~…), and the decoder generates the utterances of self-speaker (i.e., \texttt{Speaker-R}).



\bibliography{anthology,custom}

\begin{thebibliography}{42}
\expandafter\ifx\csname natexlab\endcsname\relax\def\natexlab#1{#1}\fi

\bibitem[{Bak and Oh(2019)}]{bak2019variational}
JinYeong Bak and Alice Oh. 2019.
\newblock \href {https://doi.org/10.18653/v1/D19-1202} {Variational
  hierarchical user-based conversation model}.
\newblock In \emph{Proceedings of the 2019 Conference on Empirical Methods in
  Natural Language Processing and the 9th International Joint Conference on
  Natural Language Processing (EMNLP-IJCNLP)}, pages 1941--1950, Hong Kong,
  China. Association for Computational Linguistics.

\bibitem[{Bao et~al.(2020{\natexlab{a}})Bao, He, Wang, Wu, and
  Wang}]{bao2020plato}
Siqi Bao, Huang He, Fan Wang, Hua Wu, and Haifeng Wang. 2020{\natexlab{a}}.
\newblock {PLATO}: Pre-trained dialogue generation model with discrete latent
  variable.
\newblock In \emph{Proceedings of the 58th Annual Meeting of the Association
  for Computational Linguistics}, pages 85--96.

\bibitem[{Bao et~al.(2020{\natexlab{b}})Bao, He, Wang, Wu, Wang, Wu, Guo, Liu,
  and Xu}]{bao2020plato2}
Siqi Bao, Huang He, Fan Wang, Hua Wu, Haifeng Wang, Wenquan Wu, Zhen Guo,
  Zhibin Liu, and Xinchao Xu. 2020{\natexlab{b}}.
\newblock Plato-2: Towards building an open-domain chatbot via curriculum
  learning.
\newblock \emph{arXiv preprint arXiv:2006.16779}.

\bibitem[{Cai et~al.(2020)Cai, Chen, Song, Zhang, Zhao, and
  Yin}]{cai2020Manipulation}
Hengyi Cai, Hongshen Chen, Yonghao Song, Cheng Zhang, Xiaofang Zhao, and Dawei
  Yin. 2020.
\newblock \href {https://doi.org/10.18653/v1/2020.acl-main.564} {Data
  manipulation: Towards effective instance learning for neural dialogue
  generation via learning to augment and reweight}.
\newblock In \emph{Proceedings of the 58th Annual Meeting of the Association
  for Computational Linguistics}, pages 6334--6343, Online. Association for
  Computational Linguistics.

\bibitem[{Chan et~al.(2019)Chan, Li, Yang, Chen, Hu, Zhao, and
  Yan}]{chan2019modelingwa}
Zhangming Chan, Juntao Li, Xiaopeng Yang, Xiuying Chen, Wenpeng Hu, Dongyan
  Zhao, and Rui Yan. 2019.
\newblock \href {https://doi.org/10.18653/v1/D19-1201} {Modeling
  personalization in continuous space for response generation via augmented
  {W}asserstein autoencoders}.
\newblock In \emph{Proceedings of the 2019 Conference on Empirical Methods in
  Natural Language Processing and the 9th International Joint Conference on
  Natural Language Processing (EMNLP-IJCNLP)}, pages 1931--1940, Hong Kong,
  China. Association for Computational Linguistics.

\bibitem[{Chen et~al.(2019)Chen, Peng, Wang, Xu, and Wu}]{chen2019mapping}
Chaotao Chen, Jinhua Peng, Fan Wang, Jun Xu, and Hua Wu. 2019.
\newblock \href {https://doi.org/10.24963/ijcai.2019/683} {Generating multiple
  diverse responses with multi-mapping and posterior mapping selection}.
\newblock In \emph{Proceedings of the Twenty-Eighth International Joint
  Conference on Artificial Intelligence, {IJCAI-19}}, pages 4918--4924.
  International Joint Conferences on Artificial Intelligence Organization.

\bibitem[{Chen et~al.(2017)Chen, Liu, Yin, and Tang}]{Chen2017Survey}
Hongshen Chen, Xiaorui Liu, Dawei Yin, and Jiliang Tang. 2017.
\newblock A survey on dialogue systems: Recent advances and new frontiers.
\newblock \emph{SIGKDD Explor. Newsl.}, 19(2):25–35.

\bibitem[{Chi et~al.(2017)Chi, Chen, Su, and Chen}]{chi2017speaker}
Ta-Chung Chi, Po~Chun Chen, Shang-Yu Su, and Yun-Nung Chen. 2017.
\newblock Speaker role contextual modeling for language understanding and
  dialogue policy learning.
\newblock In \emph{Proceedings of the Eighth International Joint Conference on
  Natural Language Processing (Volume 2: Short Papers)}, pages 163--168.

\bibitem[{Dai et~al.(2019)Dai, Yang, Yang, Carbonell, Le, and
  Salakhutdinov}]{dai2019transformer-xl}
Zihang Dai, Zhilin Yang, Yiming Yang, Jaime~G Carbonell, Quoc Le, and Ruslan
  Salakhutdinov. 2019.
\newblock Transformer-xl: Attentive language models beyond a fixed-length
  context.
\newblock In \emph{Proceedings of the 57th Annual Meeting of the Association
  for Computational Linguistics}, pages 2978--2988.

\bibitem[{Fleiss(1971)}]{fleiss1971measuring}
Joseph~L Fleiss. 1971.
\newblock Measuring nominal scale agreement among many raters.
\newblock \emph{Psychological bulletin}, 76(5):378.

\bibitem[{Henderson et~al.(2013)Henderson, Thomson, and
  Young}]{henderson2013deep}
Matthew Henderson, Blaise Thomson, and Steve Young. 2013.
\newblock \href {https://www.aclweb.org/anthology/W13-4073} {Deep neural
  network approach for the dialog state tracking challenge}.
\newblock In \emph{Proceedings of the {SIGDIAL} 2013 Conference}, pages
  467--471, Metz, France. Association for Computational Linguistics.

\bibitem[{{Hovy} and {Yang}(2021)}]{hovy2021SocialFactors}
Dirk {Hovy} and Diyi {Yang}. 2021.
\newblock The importance of modeling social factors of language: Theory and
  practice.
\newblock In \emph{Proceedings of the 2021 Conference of the North American
  Chapter of the Association for Computational Linguistics: Human Language
  Technologies}, pages 588--602.

\bibitem[{{Kingma} and {Ba}(2015)}]{kingma2015adam}
Diederik~P. {Kingma} and Jimmy~Lei {Ba}. 2015.
\newblock Adam: A method for stochastic optimization.
\newblock In \emph{ICLR 2015 : International Conference on Learning
  Representations 2015}.

\bibitem[{Li et~al.(2016)Li, Galley, Brockett, Spithourakis, Gao, and
  Dolan}]{li2016persona}
Jiwei Li, Michel Galley, Chris Brockett, Georgios Spithourakis, Jianfeng Gao,
  and Bill Dolan. 2016.
\newblock \href {https://doi.org/10.18653/v1/P16-1094} {A persona-based neural
  conversation model}.
\newblock In \emph{Proceedings of the 54th Annual Meeting of the Association
  for Computational Linguistics (Volume 1: Long Papers)}, pages 994--1003,
  Berlin, Germany. Association for Computational Linguistics.

\bibitem[{{Li} et~al.(2017){Li}, {Su}, {Shen}, {Li}, {Cao}, and
  {Niu}}]{li2017dailydialog}
Yanran {Li}, Hui {Su}, Xiaoyu {Shen}, Wenjie {Li}, Ziqiang {Cao}, and Shuzi
  {Niu}. 2017.
\newblock Dailydialog: A manually labelled multi-turn dialogue dataset.
\newblock In \emph{Proceedings of the Eighth International Joint Conference on
  Natural Language Processing (Volume 1: Long Papers)}, volume~1, pages
  986--995.

\bibitem[{Liu et~al.(2020{\natexlab{a}})Liu, Zhang, Zhao, Zhou, and
  Zhou}]{liu2020filling}
Longxiang Liu, Zhuosheng Zhang, Hai Zhao, Xi~Zhou, and Xiang Zhou.
  2020{\natexlab{a}}.
\newblock \href {http://arxiv.org/abs/2009.06504} {Filling the gap of
  utterance-aware and speaker-aware representation for multi-turn dialogue}.

\bibitem[{Liu et~al.(2020{\natexlab{b}})Liu, Qian, Xu, and
  Wei}]{liu2020speakerlistener}
Yafei Liu, Hongjin Qian, Hengpeng Xu, and Jinmao Wei. 2020{\natexlab{b}}.
\newblock Speaker or listener? the role of a dialogue agent.
\newblock In \emph{Proceedings of the 2020 Conference on Empirical Methods in
  Natural Language Processing: Findings}, pages 4861--4869.

\bibitem[{{Lowe} et~al.(2015){Lowe}, {Pow}, {Serban}, and
  {Pineau}}]{lowe2015Ubuntu}
Ryan {Lowe}, Nissan {Pow}, Iulian {Serban}, and Joelle {Pineau}. 2015.
\newblock The ubuntu dialogue corpus: A large dataset for research in
  unstructured multi-turn dialogue systems.
\newblock In \emph{Proceedings of the 16th Annual Meeting of the Special
  Interest Group on Discourse and Dialogue}, pages 285--294.

\bibitem[{Ma et~al.(2019)Ma, Bowden, Wu, Cui, and Walker}]{ma2019implicit}
Mingyu~Derek Ma, Kevin Bowden, Jiaqi Wu, Wen Cui, and Marilyn Walker. 2019.
\newblock Implicit discourse relation identification for open-domain dialogues.
\newblock In \emph{Proceedings of the 57th Annual Meeting of the Association
  for Computational Linguistics}, pages 666--672.

\bibitem[{Madotto et~al.(2018)Madotto, Wu, and Fung}]{madotto2018mem2seq}
Andrea Madotto, Chien-Sheng Wu, and Pascale Fung. 2018.
\newblock \href {https://doi.org/10.18653/v1/P18-1136} {{M}em2{S}eq:
  Effectively incorporating knowledge bases into end-to-end task-oriented
  dialog systems}.
\newblock In \emph{Proceedings of the 56th Annual Meeting of the Association
  for Computational Linguistics (Volume 1: Long Papers)}, pages 1468--1478,
  Melbourne, Australia. Association for Computational Linguistics.

\bibitem[{Meng et~al.(2018)Meng, Mou, and Jin}]{meng2018multiparty}
Zhao Meng, Lili Mou, and Zhi Jin. 2018.
\newblock \href {https://www.aclweb.org/anthology/L18-1496} {Towards neural
  speaker modeling in multi-party conversation: The task, dataset, and models}.
\newblock In \emph{Proceedings of the Eleventh International Conference on
  Language Resources and Evaluation ({LREC} 2018)}, Miyazaki, Japan. European
  Language Resources Association (ELRA).

\bibitem[{Mikolov et~al.(2013)Mikolov, Corrado, Kai, and
  Dean}]{Mikolov2013Word2Vec}
Tomas Mikolov, Greg Corrado, Chen Kai, and Jeffrey Dean. 2013.
\newblock Efficient estimation of word representations in vector space.
\newblock In \emph{Proceedings of the International Conference on Learning
  Representations (ICLR 2013)}.

\bibitem[{Olabiyi et~al.(2018)Olabiyi, Khazane, and
  Mueller}]{Olabiyi2018AdversarialPersona}
Oluwatobi~O. Olabiyi, Anish Khazane, and Erik~T. Mueller. 2018.
\newblock \href {https://doi.org/10.1109/ICMLA.2018.00079} {A persona-based
  multi-turn conversation model in an adversarial learning framework}.
\newblock In \emph{2018 17th IEEE International Conference on Machine Learning
  and Applications (ICMLA)}, pages 489--494.

\bibitem[{Radford et~al.(2019)Radford, Wu, Child, Luan, Amodei, and
  Sutskever}]{radford2019gpt2}
Alec Radford, Jeff Wu, Rewon Child, David Luan, Dario Amodei, and Ilya
  Sutskever. 2019.
\newblock Language models are unsupervised multitask learners.
\newblock \emph{Technical report, OpenAI}.

\bibitem[{Sankar et~al.(2019)Sankar, Subramanian, Pal, Chandar, and
  Bengio}]{sankar2019effectively}
Chinnadhurai Sankar, Sandeep Subramanian, Christopher Pal, Sarath Chandar, and
  Yoshua Bengio. 2019.
\newblock Do neural dialog systems use the conversation history effectively? an
  empirical study.
\newblock In \emph{Proceedings of the 57th Annual Meeting of the Association
  for Computational Linguistics}, pages 32--37.

\bibitem[{Serban et~al.(2016)Serban, Sordoni, Bengio, Courville, and
  Pineau}]{serban2016HRED}
Iulian~V Serban, Alessandro Sordoni, Yoshua Bengio, Aaron Courville, and Joelle
  Pineau. 2016.
\newblock Building end-to-end dialogue systems using generative hierarchical
  neural network models.
\newblock In \emph{Proceedings of the Thirtieth AAAI Conference on Artificial
  Intelligence}, pages 3776--3783.

\bibitem[{Serban et~al.(2017)Serban, Sordoni, Lowe, Charlin, Pineau, Courville,
  and Bengio}]{serban2017VHRED}
Iulian~Vlad Serban, Alessandro Sordoni, Ryan Lowe, Laurent Charlin, Joelle
  Pineau, Aaron Courville, and Yoshua Bengio. 2017.
\newblock A hierarchical latent variable encoder-decoder model for generating
  dialogues.
\newblock In \emph{Proceedings of the Thirty-First AAAI Conference on
  Artificial Intelligence}, pages 3295--3301.

\bibitem[{Shaw et~al.(2018)Shaw, Uszkoreit, and Vaswani}]{shaw2018Relative}
Peter Shaw, Jakob Uszkoreit, and Ashish Vaswani. 2018.
\newblock Self-attention with relative position representations.
\newblock In \emph{Proceedings of the 2018 Conference of the North American
  Chapter of the Association for Computational Linguistics: Human Language
  Technologies, Volume 2 (Short Papers)}, pages 464--468.

\bibitem[{Sordoni et~al.(2015)Sordoni, Bengio, Vahabi, Lioma, Grue~Simonsen,
  and Nie}]{sordoni2015HRED}
Alessandro Sordoni, Yoshua Bengio, Hossein Vahabi, Christina Lioma, Jakob
  Grue~Simonsen, and Jian-Yun Nie. 2015.
\newblock A hierarchical recurrent encoder-decoder for generative context-aware
  query suggestion.
\newblock In \emph{Proceedings of the 24th ACM International on Conference on
  Information and Knowledge Management}, pages 553--562.

\bibitem[{Sun et~al.(2021)Sun, Feng, Li, Liu, and
  Li}]{sun2021CoherentDialogueCVAE}
Bin Sun, Shaoxiong Feng, Yiwei Li, Jiamou Liu, and Kan Li. 2021.
\newblock \href {https://doi.org/10.18653/v1/2021.acl-long.437} {Generating
  relevant and coherent dialogue responses using self-separated conditional
  variational {A}uto{E}ncoders}.
\newblock In \emph{Proceedings of the 59th Annual Meeting of the Association
  for Computational Linguistics and the 11th International Joint Conference on
  Natural Language Processing (Volume 1: Long Papers)}, pages 5624--5637,
  Online. Association for Computational Linguistics.

\bibitem[{Tian et~al.(2017)Tian, Yan, Mou, Song, Feng, and
  Zhao}]{tian2017useful}
Zhiliang Tian, Rui Yan, Lili Mou, Yiping Song, Yansong Feng, and Dongyan Zhao.
  2017.
\newblock How to make context more useful? an empirical study on context-aware
  neural conversational models.
\newblock In \emph{Proceedings of the 55th Annual Meeting of the Association
  for Computational Linguistics (Volume 2: Short Papers)}, pages 231--236.

\bibitem[{Vaswani et~al.(2017)Vaswani, Shazeer, Parmar, Uszkoreit, Jones,
  Gomez, Kaiser, and Polosukhin}]{vaswani2017transformer}
Ashish Vaswani, Noam Shazeer, Niki Parmar, Jakob Uszkoreit, Llion Jones,
  Aidan~N Gomez, {\L}ukasz Kaiser, and Illia Polosukhin. 2017.
\newblock Attention is all you need.
\newblock \emph{Advances in neural information processing systems},
  30:5998--6008.

\bibitem[{Xing et~al.(2018)Xing, Wu, Wu, Zhou, Huang, and Ma}]{xing2018HRAN}
Chen Xing, Wei Wu, Yu~Wu, Ming Zhou, Yalou Huang, and Wei-Ying Ma. 2018.
\newblock Hierarchical recurrent attention network for response generation.
\newblock In \emph{Proceedings of the Thirty-Second AAAI Conference on
  Artificial Intelligence}.

\bibitem[{Xu et~al.(2021)Xu, Lei, Wang, Niu, Wu, and Che}]{xu2021DialogGraph}
Jun Xu, Zeyang Lei, Haifeng Wang, Zheng-Yu Niu, Hua Wu, and Wanxiang Che. 2021.
\newblock \href {https://doi.org/10.18653/v1/2021.acl-long.136} {Discovering
  dialog structure graph for coherent dialog generation}.
\newblock In \emph{Proceedings of the 59th Annual Meeting of the Association
  for Computational Linguistics and the 11th International Joint Conference on
  Natural Language Processing (Volume 1: Long Papers)}, pages 1726--1739,
  Online. Association for Computational Linguistics.

\bibitem[{Zhang et~al.(2019)Zhang, Lan, Pang, Guo, and Cheng}]{zhang2019ReCoSa}
Hainan Zhang, Yanyan Lan, Liang Pang, Jiafeng Guo, and Xueqi Cheng. 2019.
\newblock Recosa: Detecting the relevant contexts with self-attention for
  multi-turn dialogue generation.
\newblock In \emph{Proceedings of the 57th Annual Meeting of the Association
  for Computational Linguistics}, pages 3721--3730.

\bibitem[{{Zhang} et~al.(2018){Zhang}, {Dinan}, {Urbanek}, {Szlam}, {Kiela},
  and {Weston}}]{zhang2018personalizing}
Saizheng {Zhang}, Emily {Dinan}, Jack {Urbanek}, Arthur {Szlam}, Douwe {Kiela},
  and Jason {Weston}. 2018.
\newblock Personalizing dialogue agents: I have a dog, do you have pets too?
\newblock In \emph{Proceedings of the 56th Annual Meeting of the Association
  for Computational Linguistics (Volume 1: Long Papers)}, volume~1, pages
  2204--2213.

\bibitem[{Zhang et~al.(2018)Zhang, Cui, Wang, Zhu, Li, Zhou, and
  Liu}]{zhang2018DSHRED}
Weinan Zhang, Yiming Cui, Yifa Wang, Qingfu Zhu, Lingzhi Li, Lianqiang Zhou,
  and Ting Liu. 2018.
\newblock Context-sensitive generation of open-domain conversational responses.
\newblock In \emph{Proceedings of the 27th International Conference on
  Computational Linguistics}, pages 2437--2447.

\bibitem[{Zhang et~al.(2020)Zhang, Sun, Galley, Chen, Brockett, Gao, Gao, Liu,
  and Dolan}]{zhang2020dialogpt}
Yizhe Zhang, Siqi Sun, Michel Galley, Yen-Chun Chen, Chris Brockett, Xiang Gao,
  Jianfeng Gao, Jingjing Liu, and Bill Dolan. 2020.
\newblock {DIALOGPT} : Large-scale generative pre-training for conversational
  response generation.
\newblock In \emph{Proceedings of the 58th Annual Meeting of the Association
  for Computational Linguistics: System Demonstrations}, pages 270--278.

\bibitem[{Zhao and Eskenazi(2016)}]{zhao2016towards}
Tiancheng Zhao and Maxine Eskenazi. 2016.
\newblock \href {https://doi.org/10.18653/v1/W16-3601} {Towards end-to-end
  learning for dialog state tracking and management using deep reinforcement
  learning}.
\newblock In \emph{Proceedings of the 17th Annual Meeting of the Special
  Interest Group on Discourse and Dialogue}, pages 1--10, Los Angeles.
  Association for Computational Linguistics.

\bibitem[{Zhao and Kawahara(2019)}]{zhao2019effective}
Tianyu Zhao and Tatsuya Kawahara. 2019.
\newblock \href {http://arxiv.org/abs/1907.05599} {Effective incorporation of
  speaker information in utterance encoding in dialog}.

\bibitem[{Zhao et~al.(2020)Zhao, Xu, and Wu}]{zhao2020AuxiliaryTasks}
Yufan Zhao, Can Xu, and Wei Wu. 2020.
\newblock \href {https://doi.org/10.18653/v1/2020.emnlp-main.279} {Learning a
  simple and effective model for multi-turn response generation with auxiliary
  tasks}.
\newblock In \emph{Proceedings of the 2020 Conference on Empirical Methods in
  Natural Language Processing (EMNLP)}, pages 3472--3483, Online. Association
  for Computational Linguistics.

\bibitem[{Zheng et~al.(2020)Zheng, Yue, Huang, Chen, and
  Birch}]{Zheng2020Translation}
Zaixiang Zheng, Xiang Yue, Shujian Huang, Jiajun Chen, and Alexandra Birch.
  2020.
\newblock \href {https://doi.org/10.24963/ijcai.2020/551} {Towards making the
  most of context in neural machine translation}.
\newblock In \emph{Proceedings of the Twenty-Ninth International Joint
  Conference on Artificial Intelligence, {IJCAI-20}}, pages 3983--3989.
  International Joint Conferences on Artificial Intelligence Organization.
\newblock Main track.

\end{thebibliography}
\bibliographystyle{acl_natbib}

\appendix

\section{Input Representation Visualization}
\label{sec:appendix_input}
\begin{figure*}[t]
    \centering
    \resizebox{\textwidth}{!}{
    \includegraphics[scale=0.70]{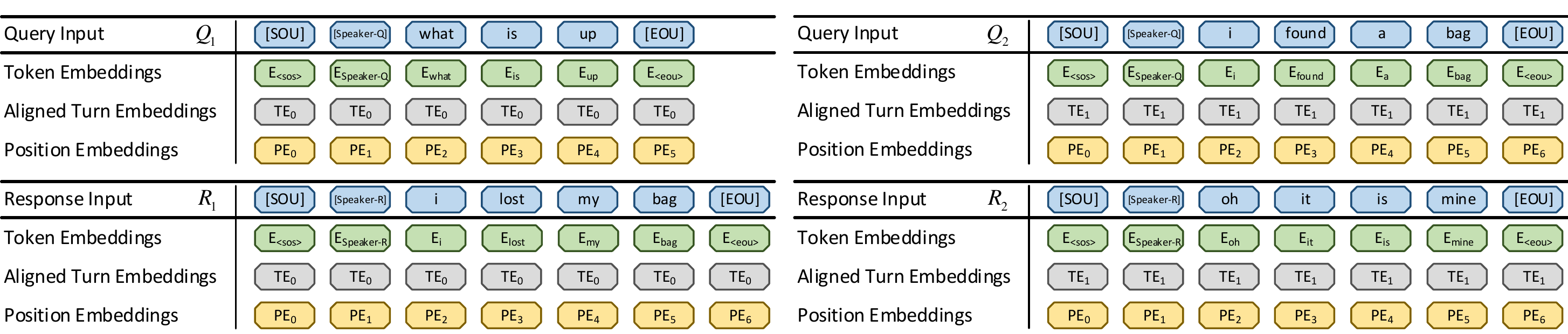}
    }
    \caption{A sample of the input representations.}
    \label{fig:input_representation_sample}
\end{figure*}
As shown in Figure~\ref{fig:input_representation_sample}, the input representation of each word is calculated by the sum of the token, aligned turn, and position feature vectors of this word.

\section{Ablation Study about $L$ on DailyDialog}
\label{sec:appendix_daily_ablation}
\begin{figure*}[t]
    \centering
    \resizebox{\textwidth}{!}{
    \subfigure{
    \includegraphics[scale=0.5]{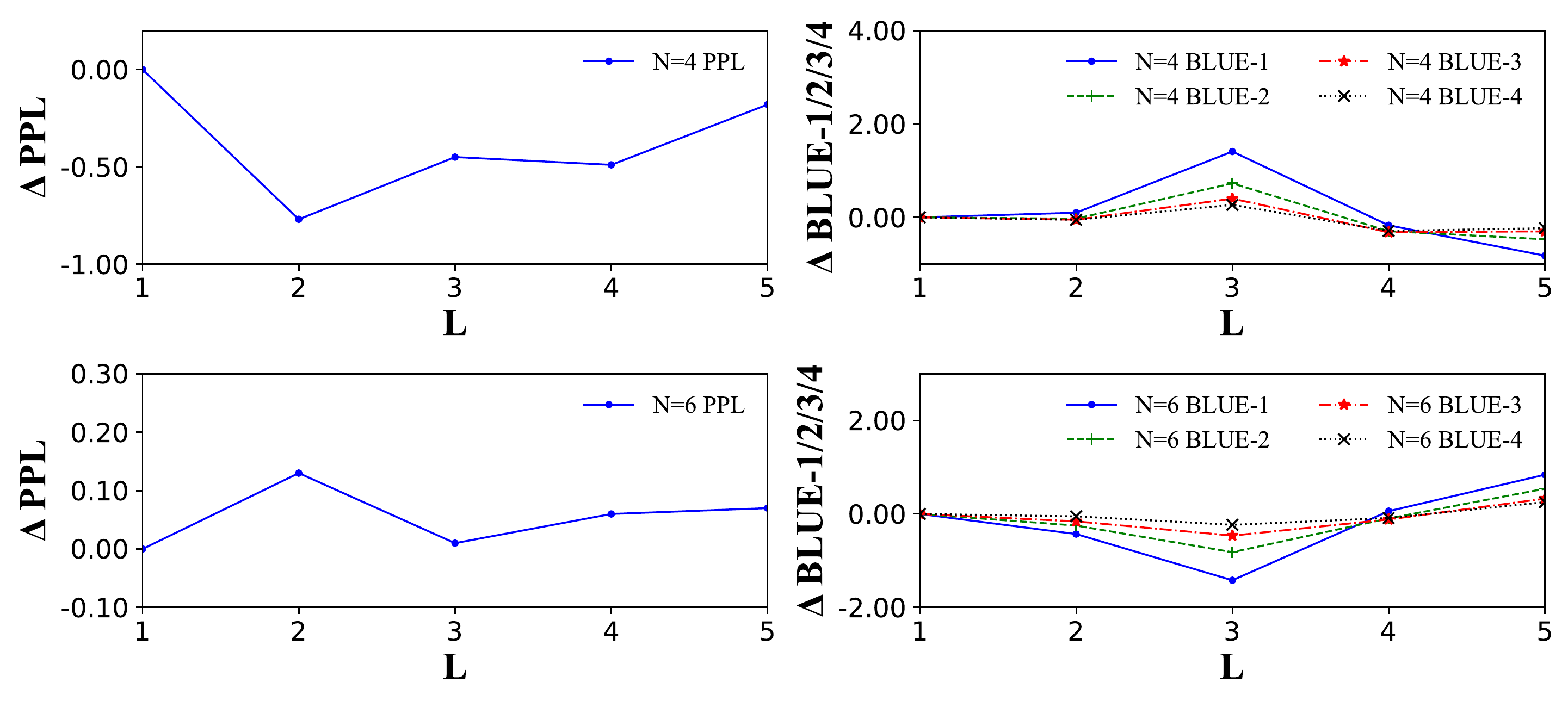}
    \label{fig:figure4a}
    \title{DailyDialog}
    }}
    \resizebox{\textwidth}{!}{
    \subfigure{
    \includegraphics[scale=0.5]{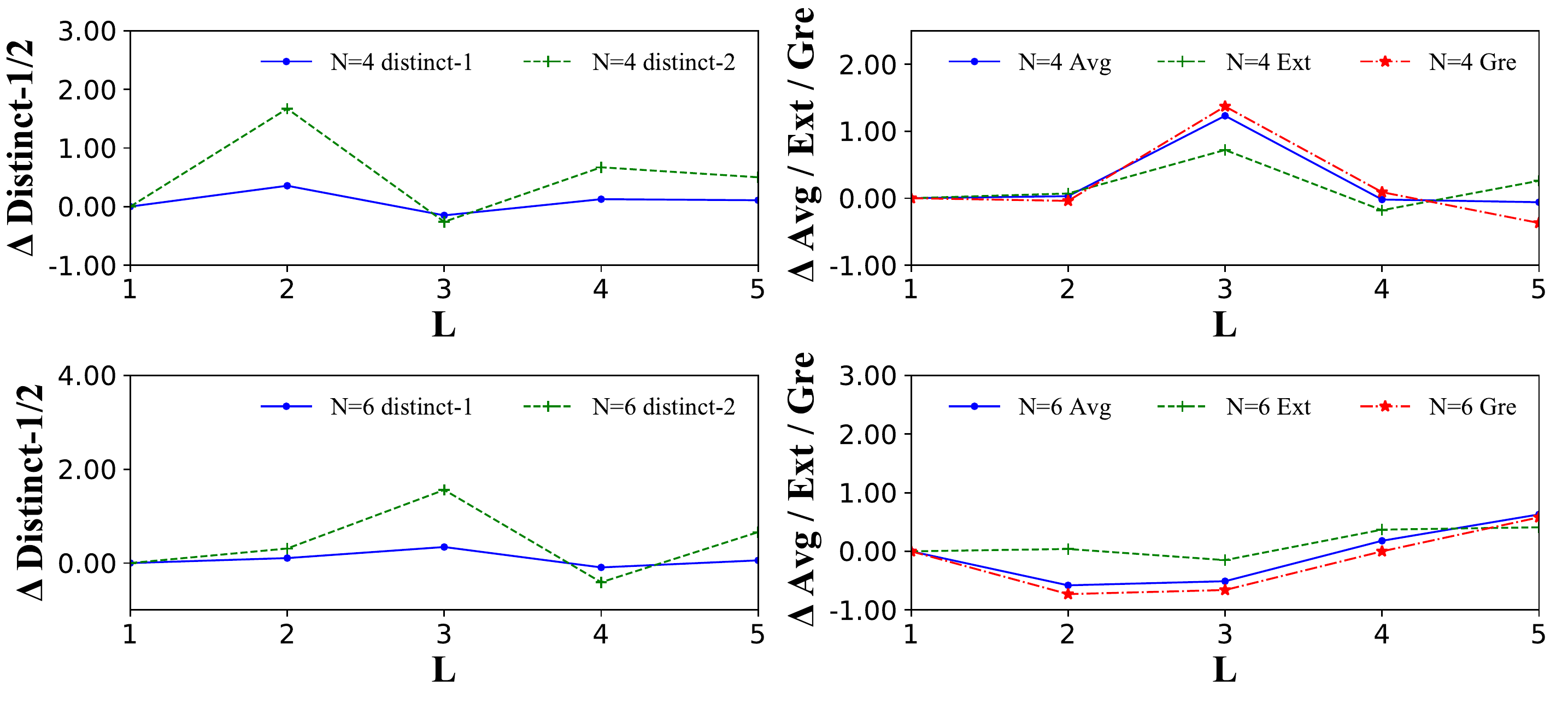}
    \label{fig:figure4b}
    \title{PersonaChat}
}}
    \caption{The impact of different memory length $L$ on the performance of PHAED on DailyDialog. The range of $L$ is from 1 to 5. }
    \label{fig:figure4_ab}
    
\end{figure*}

In the section of discussions, we re-train PHAED with different $L$ on PersonaChat to analyze the impact of memory length on the performance of PHAED. As a supplement to this analysis, we conduct the same experiment on DailyDialog and show the results in Figure~\ref{fig:figure4_ab}. Since the number of average utterances per dialogue in DailyDialog is shorter than that in PersonaChat, we set the range of $L$ from 1 to 5. We observe that the pattern of results on DailyDialog is similar to the one on PersonaChat.
With an increase of L, there are only small changes in all metrics scores, and we cannot guarantee that PHAED performs better on most metrics. Moreover, the appropriate memory length $L$ of PHAED on PersonaChat is also applicable to DailyDialog.

\section{Case Study about Two examples}
\label{sec:appendix_case_study}
In the section of discussions, we visualize the query-to-query weights of a conversation based on \emph{turn-level relative attention} of PHAED. As a supplement to the case study, we show two examples from the DailyDialog test result in Table~\ref{tab:samples_appendix}. In example 1, PHAED generates an appropriate and informative response, but other baselines either generate responses from the wrong speaker perspective or generate short and safe responses. In example 2, PHAED generates a response that includes clear location information. However, DialGPT and ReCoSa generate responses based on a wrong previous query, and the other responses only contain fuzzy location information. 

\begin{table}[t]
\centering
\resizebox{\columnwidth}{!}{
\begin{tabular}{ll}
    \textbf{Sample 1} & \\
    \hline
    \textbf{History} & 
    \makecell[l]{
    Speaker-R May i help you? \\
    Speaker-Q I 'm looking for a pair of shoes. \\
    \makecell[l]{Speaker-R What size do you wear? these shoes are on sale.} \\
    \makecell[l]{Speaker-Q I wear size 36. this pair looks nice. i try them on?}}\\
    \hline 
    HRAN & Speaker-R Of course.\\
    DSHRED & Speaker-R Of course. it's a bit too expensive.\\
    SpkHRED & Speaker-R Of course. you can have a look at the moment.\\
    Transformer & Speaker-R Yes. they look nice. how much is it? \\
    ReCoSa & Speaker-R Yes, they are. \\
    DialGPT & Speaker-R Yes, of course. \\
    \textbf{PHAED} & Speaker-R Of course. they are all handmade.\\
    \hline
    \textbf{Sample 2} & \\
    \hline
    \textbf{History} & 
    \makecell[l]{Speaker-S 911. how may i help you? \\
    Speaker-R I need to report a break in. \\
    Speaker-S When did this break in occur? \\
    Speaker-R Last night. \\
    Speaker-S Where did the break in happen?}\\
    \hline 
    HRAN & Speaker-R I was in the middle of the city.\\
    DSHRED & Speaker-R It was in the middle of the city.\\
    SpkHRED & Speaker-R Well, i'm sorry, sir. we have a nice room.\\
    Transformer & Speaker-R It was a very good game.\\
    ReCoSa & Speaker-R It was in the middle of October.\\
    DialGPT &  Speaker-R It's a couple of days ago.\\
    \textbf{PHAED} & Speaker-R It happened at the hilton hotel.\\
    \hline
\end{tabular}}
\vspace{-2ex}
\caption{Two samples from the DailyDialog test result.}
\vspace{-3ex}
\label{tab:samples_appendix}
\end{table}

\end{document}